\documentclass[a4paper, 10pt, conference]{ieeeconf}
\IEEEoverridecommandlockouts                              
                                                          
\usepackage{amsmath,amsfonts}
\usepackage{algorithmic}
\usepackage{algorithm}
\usepackage{array}
\usepackage[caption=false,font=normalsize,labelfont=sf,textfont=sf]{subfig}
\usepackage{textcomp}
\usepackage{stfloats}
\usepackage{url}
\usepackage{verbatim}
\usepackage{graphicx}
\usepackage{cite}
\usepackage{threeparttable }
\usepackage{amsmath}
\usepackage{amssymb} 
\newcommand{\sign}{\text{sgn}}

\usepackage[pscoord]{eso-pic}

\newcommand{\placetextbox}[3]{
  \setbox0=\hbox{#3}
  \AddToShipoutPictureFG*{
    \put(\LenToUnit{#1\paperwidth},\LenToUnit{#2\paperheight}){\vtop{{\null}\makebox[0pt][c]{#3}}}%
  }%
}%

\begin{document}
\placetextbox{0.5}{0.94}{\texttt{This work has been submitted to the IEEE for possible publication. Copyright may be}}%
\placetextbox{0.5}{0.92}{\texttt{transferred without notice, after which this version may no longer be accessible.}}%

\title{\LARGE \bf  Design and Characterization of Viscoelastic McKibben Actuators with Tunable Force-Velocity Curves}



\author{Michael~J.~Bennington$^\text{* [1]}$,
      Tuo~Wang$^\text{* [1]}$,
      Jiaguo~Yin$^\text{[2]}$, \\
      Sarah~Bergbreiter$^\text{[1]}$, Carmel~Majidi$^\text{[1]}$, Victoria~A.~Webster-Wood$^\text{+ [1,3,4]}$ 
\thanks{*: The authors contributed equally. +: Corresponding author {\tt\small vwebster@andrew.cmu.edu}. Departments of [1] Mechanical Engineering, [2] Materials Science and Engineering, [3] Biomedical Engineering. [4] McGowan Institute for Regenerative Medicine. All departments and institutes are part of Carnegie Mellon University, 5000 Forbes Ave, Pittsburgh, PA, 15213, USA.}}




\maketitle

\begin{abstract}
The McKibben pneumatic artificial muscle is a commonly studied soft robotic actuator, and its quasistatic force-length properties have been well characterized and modeled. However, its damping and force-velocity properties are less well studied. Understanding these properties will allow for more robust dynamic modeling of soft robotic systems. The force-velocity response of these actuators is of particular interest because these actuators are often used as hardware models of skeletal muscles for bioinspired robots, and this force-velocity relationship is fundamental to muscle physiology. In this work, we investigated the force-velocity response of McKibben actuators and the ability to tune this response through the use of viscoelastic polymer sheaths. These viscoelastic McKibben actuators (VMAs) were characterized using iso-velocity experiments inspired by skeletal muscle physiology tests. A simplified 1D model of the actuators was developed to connect the shape of the force-velocity curve to the material parameters of the actuator and sheaths. Using these viscoelastic materials, we were able to modulate the shape and magnitude of the actuators' force-velocity curves, and using the developed model, these changes were connected back to the material properties of the sheaths. 
\end{abstract}

\section{Introduction}
Originally introduced in the 1930s-1940s\cite{daerden2002pneumatic}, and popularized by Joseph McKibben in the 1950s \cite{Hawkesetal2021Questions,Tondu2012,Chou1996}, pneumatic artificial muscles are a commonly studied soft robotic actuator and have been used in traditional rigid robotics \cite{daerden2002pneumatic,Delson2005,Kurumaya2016}, soft robotic platforms \cite{controlMcKibbenDesign,Faudzi2018}, and wearable and assistive devices \cite{Connollyetal2016trajectory,Tschiersky2020,Koizumi2020,Rosaliaetal2022sleeve,Parketal2012}. Consisting of an inner rubber bladder and an outer constraining mesh, the McKibben actuator is able to achieve high actuator strains and large force relative to its light weight \cite{Tondu2006}. McKibben actuators are of particular interest in bioinspired robotics and prosthetics because of their functional similarity to biological muscle in terms of contracting in response to activation and introducing compliance into the system. As a consequence, they can serve as first-order hardware models of skeletal muscle \cite{Klute1999,Chou1996,Tondu2006,Gollobetal2022Joint}. Current experimental characterizations and models of these actuators tend to focus on their quasistatic properties, relating their inflation pressure, length, and axial force \cite{Tondu2012,Al-Ibadi2017,Olsenetal2022McKibbenModeling,Kothera2009, Taimooretal2019}, but less attention has been given to their dynamics properties. 
These properties are important both to the design and modeling of the dynamics of a robotic system composed by these actuators and to the use of McKibben muscles as biomimetic actuators.

\begin{figure}[t]
    \centering
    \includegraphics[width=0.7\linewidth]{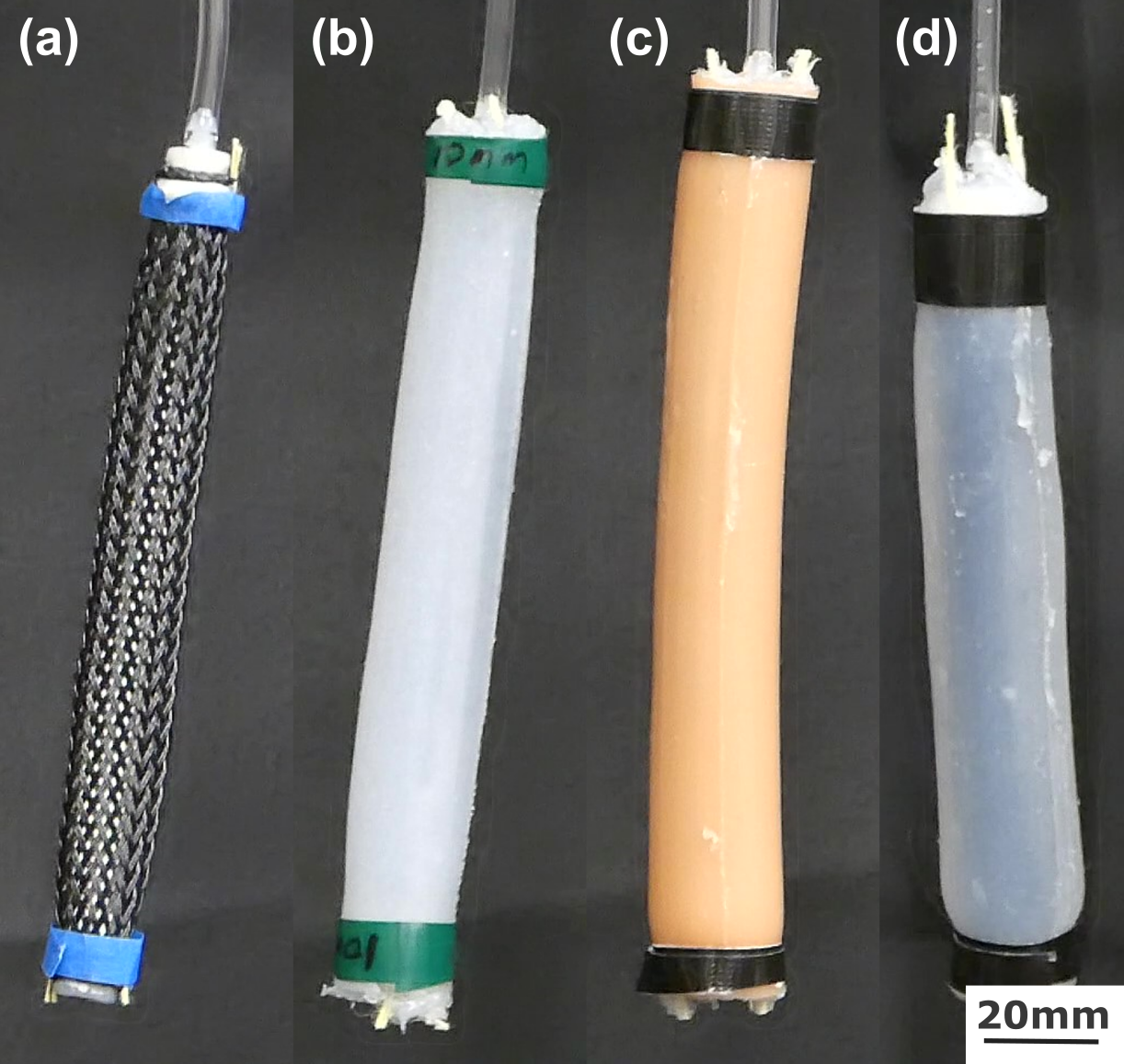}
    \caption{Viscoelastic McKibben Actuator (VMA): (a) Plain McKibben Actuator (control), (b) Ecoflex-30 sheath, (c) Urethane sheath, (d) Ecoflex-30 and Carbopol composite sheath (10mm diameter shown for all). Each VMA contains a plain McKibben actuator at its core, fabricated in the same method as the control.}
    \label{fig:Intro}
    \vspace{-15 pt}
\end{figure}

While few studies have been reported on the dynamic properties of McKibben actuators, those that have done so have often focused on the force-velocity relationship. For example, Tondu et al. performed isotonic quick-release experiments on McKibben actuators and showed that, for a particular combination of rubber bladder and mesh materials, the force-velocity relationship can resemble that of the Hill muscle model \cite{Tondu2006}. Other works have shown that the velocity-dependence of the McKibben actuator's force is minimal compared to that of biological muscle\cite{Klute1999,Chou1994}. The authors instead augmented the muscle with parallel hydraulic damping elements to better mimic the biological tissue \cite{Klute1999}. However, these solutions either rely on very particular woven mesh materials or large auxiliary equipment to tune the shape of the actuator's force-velocity response. 

In this work, we begin to investigate the force-velocity relationship of McKibben actuators and the ability to tune these relationships using simple viscoelastic material sheaths. Four different actuator architectures are investigated using actuators of three different diameters. The force-velocity response of these viscoelastic McKibben actuators (VMA) is measured using iso-velocity tests adapted from the muscle physiology literature \cite{Yu1999}. To connect the measured force-velocity response to the material properties of the sheath and the mechanics of the underlying McKibben actuator, a simplified 1D model, consisting of parallel chains of standard linear solid elements (SLSEs), is formulated. 


\section{Materials and Methods}

\begin{figure*}[t]
    \centering
    \includegraphics[width=\linewidth]{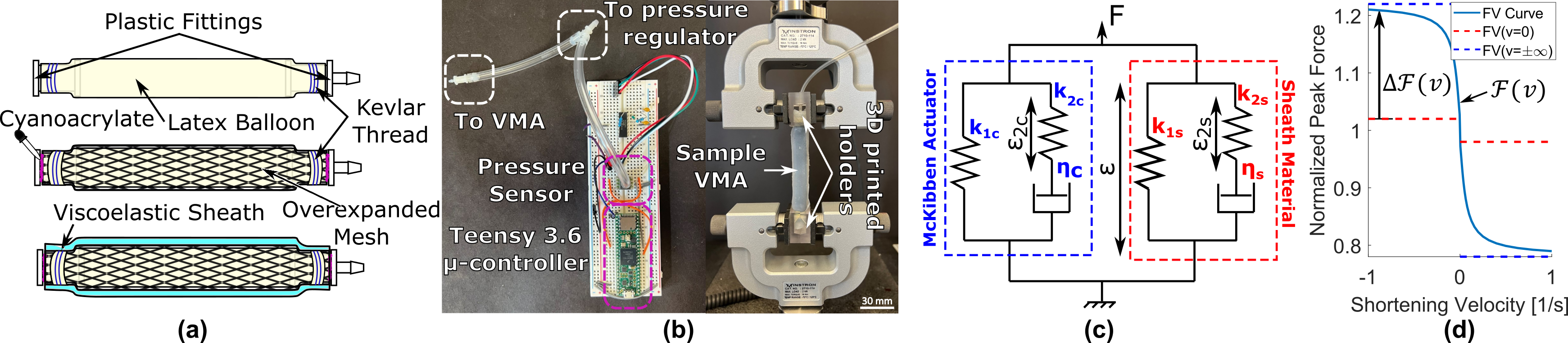}
    \caption{Fabrication, Characterization, and Modeling. (a) Each viscoelastic muscle actuator consists of a standard McKibben actuator (fabricated following \cite{controlMcKibbenDesign}) and a viscoelastic polymer sheath. (b) To characterize the dynamic properties of the actuators, iso-velocity experiments were performed on an Instron 5969 at various velocities and inflation pressures. (c) The dynamics of the actuators were modeled using parallel chains of Standard Linear Solid elements (SLSE), with one arm capturing the dynamics of the McKibben actuator and the other the dynamics of the sheath material. Using this model, an analytical expression for the force-velocity curves can be obtained (d), and the shape of the curve can be related to the material properties of the constituents. The height of this curve above the $v=0$ point, $\Delta FV(v)$, can be related to two material properties of the actuator. Here, shortening velocity (negative of the extension rate) is reported in alignment with standard muscle physiology experiments.}
    \label{fig:VMA_and_setup}
    \vspace{-15 pt}
\end{figure*}

\subsection{Actuator Design and Fabrication}

Each viscoelastic muscle actuator consists of a traditional McKibben actuator, serving as the contractile element, and a viscoelastic sheath around the McKibben, serving as a passive damper (Fig. \ref{fig:VMA_and_setup}a). Four 90 mm long McKibben actuators each of three different diameters (6 mm, 10 mm, 12 mm nominal mesh diameter) were fabricated. The design of the actuator was adapted from \cite{controlMcKibbenDesign}. Briefly, a latex balloon inner bladder is connected to two barbed tube ends and is constrained by commercially available overexpanded cable meshes (PET Expandable Sleeving, Alex Tech). Kevlar fibers and cyanoacrylate glue were used to seal and connect the bladder and mesh to the end caps of the actuator.




Thin hollow sheaths of different viscoelastic and thixotropic materials were attached to the outside of the McKibben to act as the damping element of the actuator. To create the outer viscoelastic sheaths, 2 single-layered, concentric-cylindrical molds were 3D printed (Object 30, Stratasys), with inner diameters of 9 mm and 12 mm. For both diameters, the resulting sheath has a thickness of 2 mm. Polyurethane (Vytaflex, Smooth-On Inc.), Ecoflex-30 (Ecoflex 00-30, Smooth-On Inc.) and 5\% Carbopol (Carbomer 940, Sanare) gel were used to fabricate the McKibben sheaths. For both the Ecoflex-30 and polyurethane sheaths, the liquid elastomer was prepared by mixing the 2-part polymer in a 1:1 ratio. The mixed polymer was placed in a vacuum chamber for 5 minutes to remove air bubbles. The 3D-printed molds were prepared by spraying a thin layer of mold release (Ease Release 200, Mann Release Technologies) on the inner surfaces of the mold. The elastomer was then injected into the mold and cured at room temperature (25$^\circ$C) for 12 hours. The Carbopol gel used in this project was adapted from \cite{Carbopol1}. First, 10g of Carbopol 940 powder (Carbomer) was mixed with 190g of deionized water. The mixture was then mechanically stirred for 4 hours. After stirring, 4g of 10M NaOH solution was added to the mixture. The new mixture was then mechanically stirred for 30 min. Finally, the gel was injected in between an Ecoflex-30 sheath and the McKibben actuator. The resulting sheaths were connected at the ends of the actuator using silicone epoxy (Sil-poxy, Smooth-On Inc.) and Kevlar threads. 

The geometric and material parameters of all 12 actuators fabricated for experimental characterization, with and without sheaths, are provided in Table~\ref{McKibbenParameters}. The fabricated length of each actuator was measured by a digital caliper. The Max Contraction Ratio is defined as the ratio of the length of the actuator at 20 psi to the length of the actuator at 0 psi (initial length).

\subsection{Experimental Characterization}

Inspired by biological muscle testing \cite{Yu1999}, iso-velocity tests were performed at different pressure levels for all sample actuators on a universal material testing system (5969, Instron, 1 kN load cell). Inflation pressure was measured with a digital pressure sensor (ELVH-030G-HAND-C-PSA4, ALL SENSORS, maximum pressure 30 psi, resolution 0.1 psi) and recorded using a microcontroller (Teensy 3.6, PJRC). Two pairs of 3D printed holders were designed to hold both ends of the actuator and provide consistent friction between the actuator and the testing system. The force and length data from the universal material testing system and pressure data from the microcontroller were collected independently and synchronized later in MATLAB.

\begin{figure}[t]
    \centering
    \includegraphics[width=0.95\linewidth]{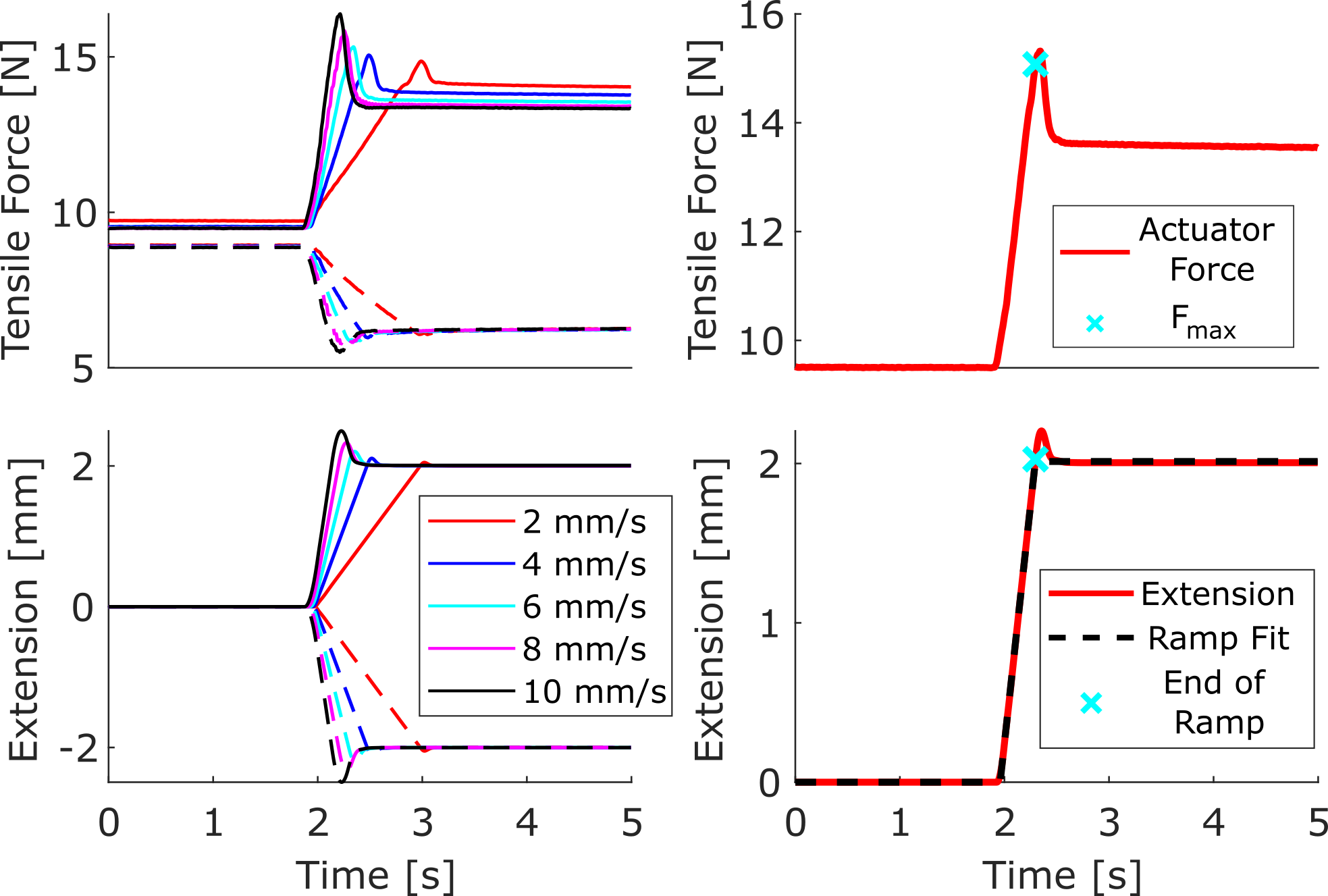}
    \caption{Characterization and Modeling of the Force-Velocity Curve. (a) An example iso-velocity experiment (6mm control McKibben actuator at 10 PSI) with the individual velocity ramps overlayed. Data from these force responses are used to construct an experimental force-velocity curve. (b) To avoid confounding effects from various amounts of overshoot, the peak force that is normalized by the initial force and the velocity is normalized by the pressurized rest length for the force-velocity curve is taken at the point when the ramp first reaches its target point. This occurs just prior to the extension overshoot.}
    \label{fig:Test_Profile}
    \vspace{-15 pt}
\end{figure}

Iso-velocity tests (Fig.~\ref{fig:Test_Profile} (a)) were performed at five velocity magnitudes (2, 4, 6, 8, 10, all in mm/s) at 4 pressure levels (5, 10, 15, 20, all in psi). All five velocities were tested in a single session at a given pressure level. For a given pressure: the actuator was first held at its rest length in the testing system and pressurized to the desired level. After allowing the actuator force to reach steady state, the actuator was stretched between +4 and -4 mm at 0.01 mm/s for one cycle,  returned to the unpressurized rest length, and again allowed to come to steady state. This was done to minimize preconditioning effects on the first ramp. For each velocity magnitude $v$: the actuator was stretched 2 mm at $v$ mm/s and then held for 30 seconds. The actuator was then returned to the unpressurized rest length at 0.01 mm/s and held for 30 seconds. This same profile was then repeated at a velocity of $-v$ mm/s. For 5 psi, only 1 mm of extension was applied, as shortening by more than 1mm from the \textit{unpressurized} rest length would have led to shortening below the \textit{pressurized} rest length of the actuators. Five repetitions of this full protocol were conducted for each actuator at each pressure and velocity.

\begin{table*}[!hb]
\caption{Geometric and Material Parameters for the Viscoelastic McKibben Actuators}
\scalebox{1}{
    \centering
    \label{McKibbenParameters}
    \begin{tabular}{p{0.05\linewidth}p{0.1\linewidth}p{0.125\linewidth}p{0.14\linewidth}p{0.12\linewidth}p{0.14\linewidth}p{0.146\linewidth}}
    \hline
    Sample & Mesh Diameter (mm) & IL* $\pm$ 1 STD (mm) & ML*  $\pm$ 1 STD (mm) & Max Contraction Ratio (\%) & Sheath Material & Sheath Diameter (mm) \\
    \hline 
    \hline
    Control1 & 6 & 88.5$\pm$0.9 & 68.3$\pm$0.6 & 22.8 & N/A & N/A\\
    Control2 & 10 & 94.3$\pm$0.5 & 70.3$\pm$0.3 & 25.4 & N/A & N/A\\
    Control3 & 12 & 91.5$\pm$0.4 & 67.2$\pm$0.1 & 26.6 & N/A & N/A\\
    Ecoflex1 & 6 & 91.7$\pm$0.2 & 75.2$\pm$0.3 & 17.9 & Ecoflex-30 & 9\\
    Ecoflex2 & 10 & 89.7$\pm$0.4 & 70.2$\pm$0.5 & 21.8 & Ecoflex-30 & 9\\
    Ecoflex3 & 12 & 90.9$\pm$0.6 & 69.6$\pm$0.5 & 23.4 & Ecoflex-30 & 12\\
    Urethane1 & 6 & 91.2$\pm$1.0 & 70.6$\pm$0.5 & 21.7 & Poly-urethane & 9\\
    Urethane2 & 10 & 90.8$\pm$0.7 & 70.2$\pm$0.6 & 22.6 & Poly-urethane & 12\\
    Urethane3 & 12 & 89.7$\pm$0.4 & 68.3$\pm$0.8 & 23.9 & Poly-urethane & 12\\
    Carbopol1 & 6 & 92.5$\pm$0.3 & 72.7$\pm$0.3 & 21.4 & Carbopol+Ecoflex-30 & 12\\
    Carbopol2 & 10 & 93.8$\pm$0.4 & 70.2$\pm$0.5 & 21.9 & Carbopol+Ecoflex-30 & 12\\
    Carbopol3 & 12 & 86.4$\pm$0.3 & 64.8$\pm$0.4 & 24.9 & Carbopol+Ecoflex-30 & 12\\
    \end{tabular}}
    \begin{tablenotes}
    \item IL*: Initial length. The length of sample actuator measured at 0 psi.
    \item ML*: Minimum length. The length of sample actuators measured at 20 psi.
    \item STD: Standard Deviation.
    
    \end{tablenotes}
\end{table*}
\begin{figure}[th]
    \centering
    \includegraphics[width=\linewidth]{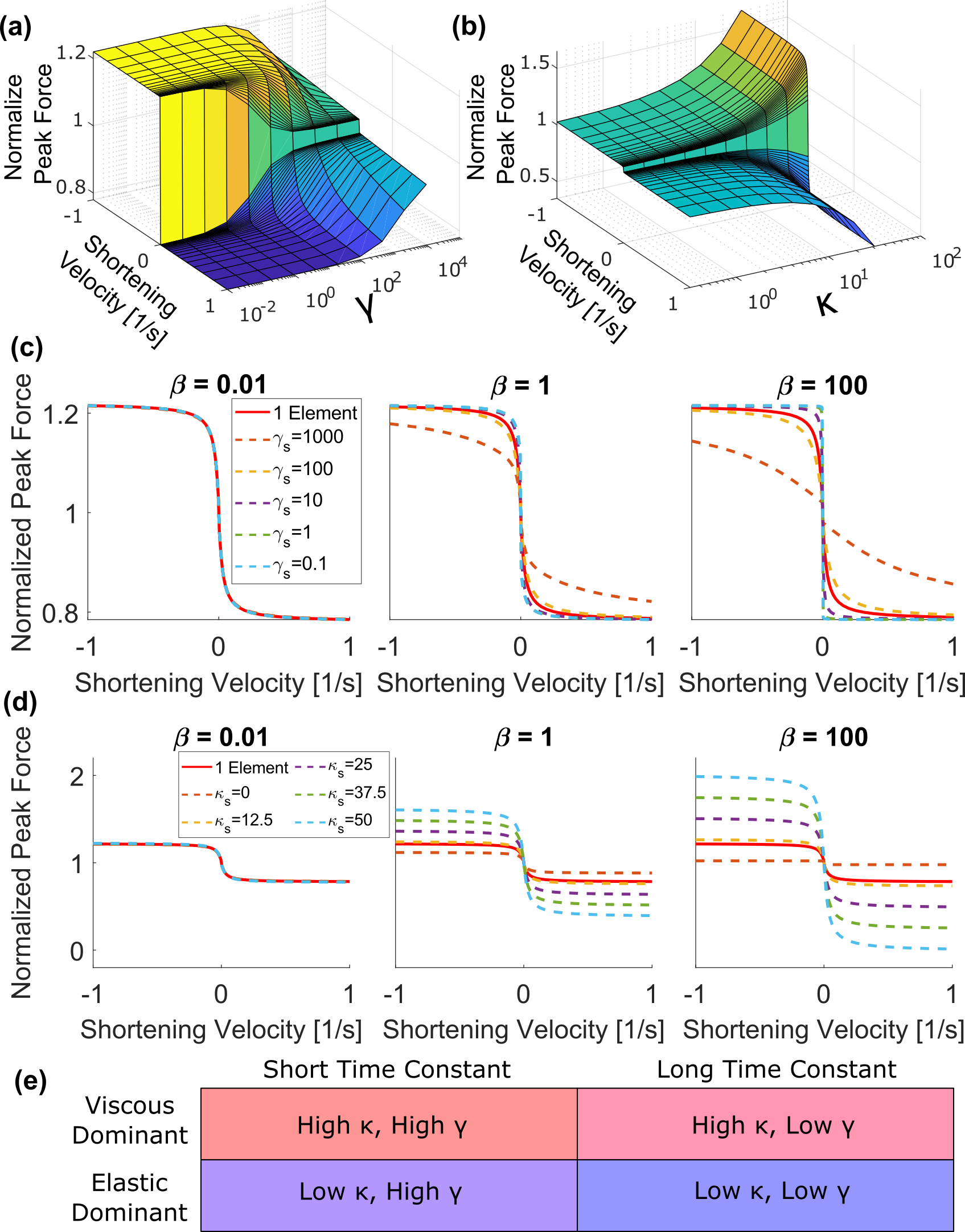}
    \caption{Investigation of model parameters for a 1-SLSE model ((a) and (b)) and for a 2-SLSE model ((c) and (d)). Here the normalized shortening velocity (negative of the extension strain rate, $v$) is reported in alignment with standard muscle physiology experiments. (a) By varying the stiffness : damping ratio in the viscous arm of the SLSE, the slope of the force-velocity curve can be changed. As $\gamma$ decreases (increased damping time constant), the force-velocity curve approaches a step response, with no velocity dependence. Conversely, as $\gamma$ increases, the curve approaches a linear response. (b) By varying the stiffness ratio between the two arms of the model, the height of the force-velocity curve is changed, with the height increasing with increasing $\kappa$. For (c) and (d), one SLSE in the model was fixed with $\kappa_1=10$ and $\gamma_1=50$, and the parameters of the other arm were varied. (c) By varying $\gamma_2$ in the second arm, the slope of the force-velocity can be tuned, and (d) by varying $\kappa_2$, the height of the force-velocity curve can be adjusted. In both cases, increasing $\beta$ (the stiffness ratio of the two parallel elastic elements), the effect of changing $\kappa_2$ or $\gamma_2$ is amplified. (e) These parameters can grouped into four different material classes. By combining materials of classes, the force-velocity curve can be tuned for a desired dynamic response.}
    \label{fig:Parameter_Sweep}
    \vspace{-15 pt}
\end{figure}

\subsection{Modeling}
To relate changes in the experimental force-velocity curves to design parameters of the actuator materials, a simplified, 1D model was developed where both the McKibben actuator and polymer sheaths were treated as standard linear solid elements (SLSE) (Fig. \ref{fig:VMA_and_setup}c). The resulting force-velocity expressions are parameterized by mechanical properties of the actuator constituents and can therefore be used as a design tool to inform future designs. In this model, elastic elements are assumed to have a force linearly proportional to strain ($F = k \varepsilon$ where the normalized stiffness $k$ has units [N]), and the viscous elements are assumed to have a force linearly proportional to the strain rate ($F = \eta \dot\varepsilon$ where the damping coefficient $\eta$ has units [Ns]). For the case of a single SLSE, the system force is given by:
\vspace{-5 pt}
\begin{equation}
    F(t) = k_{1_i} \varepsilon(t) + k_{2_i} \varepsilon_{2_i}(t)
    \label{eqn:systemforce}
\end{equation}
where $k_{1_i}$ and $k_{2_i}$ are the stiffness of the parallel and series elastic elements, and $\varepsilon$ and $\varepsilon_{2_i}$ are the strains of the parallel and series elastic elements of the $i^\text{th}$ SLSE. For the actuators presented here only two SLSEs are included: a control McKibben (c) and the sheath (s). For each SLSE,
\vspace{-5 pt}
\begin{equation}
    \dot\varepsilon_{2_i} = \dot\varepsilon - \frac{k_{2_i}}{\eta_i} \varepsilon_{2_i}
    \label{eqn:damping}
\end{equation}
where $\eta_i$ is the damping coefficient of the series damper. Starting from steady state ($\dot\varepsilon_{2_i}(t=0^-) =0$), a constant strain rate ramp ($\dot\varepsilon=\hat v$ where $\hat v$ has units [1/s]) yields a system force
\vspace{-5 pt}
\begin{equation}
    F_i(t) = k_{1_i} (\varepsilon_0 + \hat vt) + \frac{\hat v \eta_i}{k_{2_i}}(1 - e^{-\frac{k_{2_i}}{\eta_i} t}) \,.
    \label{eqn:peaksystemforce}
\end{equation}
For a fixed final applied strain $d\varepsilon$, the peak system force is a function of the velocity ($t_{peak} = d\varepsilon/\hat v$). Normalizing by the pre-extension, steady state force ($F_i(t=0^-) = k_{1_i} \varepsilon_0$), the force-velocity curve for the model can be written as:
\vspace{-5 pt}
\begin{equation}
    FV_i(\hat v) = 1 + \sign(\hat v)\frac{d\varepsilon}{\varepsilon_0} + \frac{\hat v \kappa_i}{\varepsilon_0 \gamma_i}(1-e^{-\sign(\hat v)\frac{\gamma_i d\varepsilon}{\hat v}})
    \label{eqn:relativestiffness}
\end{equation}
where $\kappa_i = k_{2_i} / k_{1_i}$ is the relative stiffness of the elastic elements and $\gamma_i = k_{2_i} / \eta_i$ is the inverse of the time constant of the viscous arm (Fig. \ref{fig:VMA_and_setup}d). Here, $\sign(x)$ is the sign function ($1: x>0, -1: x<0$). The height of the force-velocity curve above the $v=0$ discontinuity then takes the form:
\vspace{-5 pt}
\begin{equation}
    \Delta FV_i(\hat v) = \frac{\hat v \kappa_i}{\varepsilon_0 \gamma_i}(1-e^{-\sign(\hat v)\frac{\gamma_i d\varepsilon}{\hat v}}) \,.
    \label{eqn:FVheight}
\end{equation}
Using this equation, the parameters $\kappa_i$ and $\gamma_i$ can be related to the shape of force-velocity curve. Specifically, the horizontal asymptote is given by:
\vspace{-5 pt}
\begin{equation}
    \Delta FV_i(\hat v_\infty) = \frac{d\varepsilon}{\varepsilon_0} \kappa_i
    \label{eqn:horizontalasymptote}
\end{equation}
and the velocity $\hat v_\alpha$ at which the force-velocity curve reaches $\alpha \Delta FV_i(\hat v_\infty)$ can be approximated as:
\vspace{-5 pt}
\begin{equation}
    \hat v_\alpha \approx \frac{d\varepsilon}{2(1-\alpha)} \gamma_i \,.
    \label{eqn:peakvelocity}
\end{equation}
This approximation is valid within 5\% for $\alpha > 0.75$. Thus the height of the force-velocity curve is governed by $\kappa_i$ and the steepness of the force-velocity curve is governed by $\gamma_i$ (Fig.\ref{fig:Parameter_Sweep}a,b).

For the two-SLSE case (McKibben actuator and the sheath), similar relationships can be found. The shape of the force-velocity curve takes the form:
\vspace{-5 pt}
\begin{equation}
    \Delta FV_{c+s}(\hat v) = \frac{\beta_c}{\beta_c + \beta_s} \Delta FV_{c}(\hat v) + \frac{\beta_s}{\beta_c + \beta_s} \Delta FV_{s}(\hat v)
    \label{eqn:FVshapes}
\end{equation}
where $\Delta FV_s$ and $\Delta FV_c$ both take the form of $\Delta FV_i$ from the 1 SLSE case. The height and steepness are governed by weighted averages of $\kappa_i$ and $\gamma_i$:
\vspace{-5 pt}
\begin{equation}
    \Delta FV_{c+s}(\hat v_\infty) = \frac{d\varepsilon}{\varepsilon_0} \frac{\beta_c \kappa_c + \beta_s \kappa_s}{\beta_c + \beta_s}
    \label{eqn:FVheightandsteepness1}
\end{equation}
and 
\vspace{-5 pt}
\begin{equation}
    \hat v_\alpha \approx \frac{d\varepsilon}{2(1-\alpha)} \frac{\beta_c \kappa_c \gamma_c + \beta_s \kappa_s \gamma_s}{\beta_c \kappa_c + \beta_s \kappa_s}
    \label{eqn:FVheightandsteepness2}
\end{equation}
where $\beta_i = k_{1_i} / k_{1_c}$ is the stiffness ratio of parallel elastic element to the McKibben parallel stiffness ($\beta_c = 1$) in the different elements. With different combinations of $\beta_s$, $\kappa_s$, and $\gamma_s$, the force-velocity response of the McKibben actuator can be tuned (Fig. \ref{fig:Parameter_Sweep} (c),(d)). These expressions can also be extended to any number of parallel SLSEs following this weighted average scheme.

\begin{figure*}[t]
    \centering
    \includegraphics[width=0.85\linewidth]{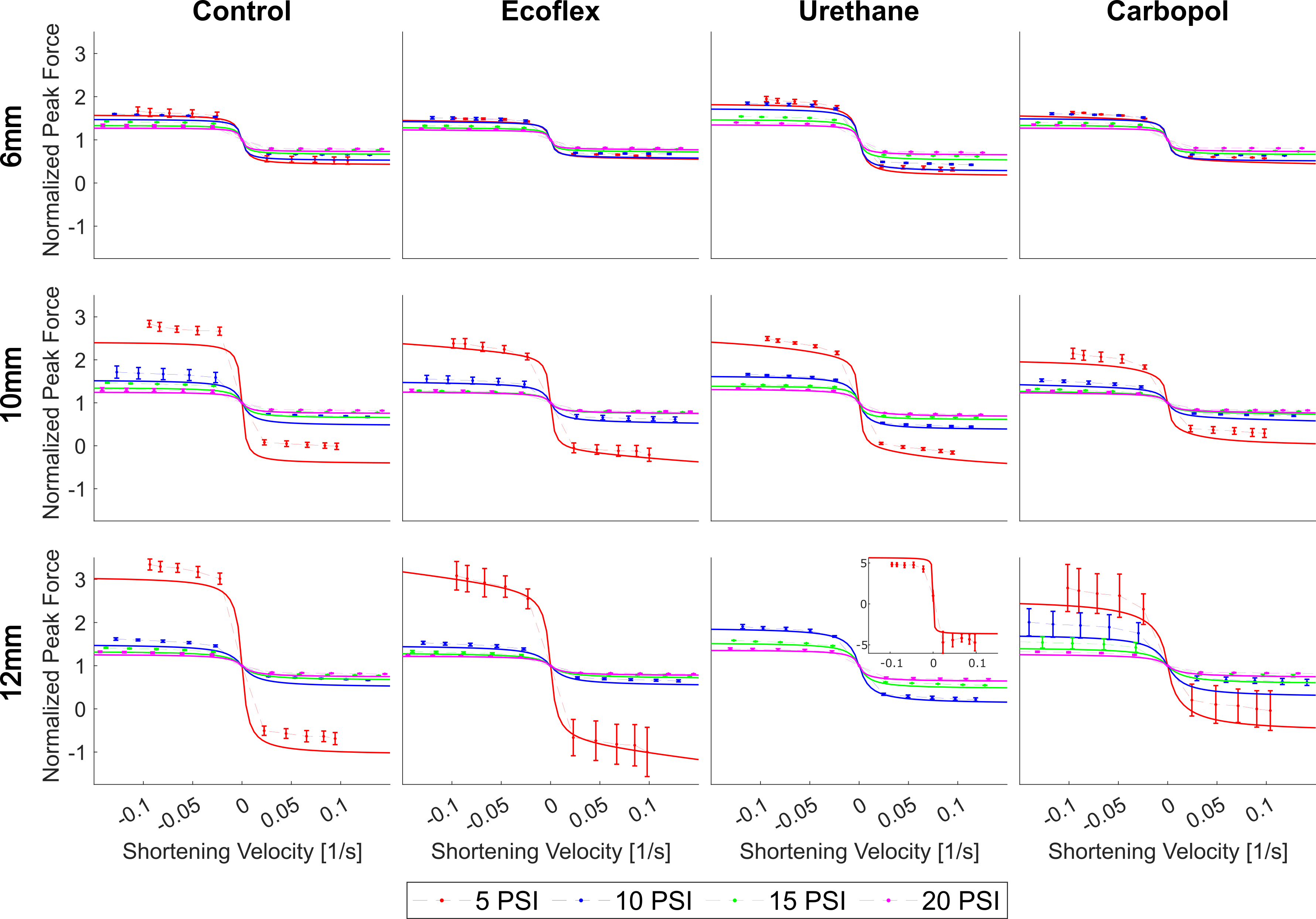}
    \caption{Experimental Characterization of Viscoelastic McKibben Actuators. Each column shows the data for a different actuator, and each row shows a different actuator diameter. Along the dashed line, the experimental data is reported as mean $\pm$ 1 standard deviation. The solid line shows the corresponding model fit for that actuator and that pressure. For the control actuators, a 1-SLSE model is used, and for each of the VMA, a 2-SLSE model is used, with the control element parameters set by the corresponding control actuator model. Inset: The 5PSI curve for the 12mm Urethane actuator is inset to allow a smaller axis range for the rest of the 12mm actuators.  }
    \label{fig:Experimental_Data}
    \vspace{-15 pt}
\end{figure*}

\subsection{Analysis}
Experimental force-velocity curves were compiled for each actuator and each pressure level using data measured during characterization experiments. Individual velocity ramps were identified and extracted from the larger experiment (Fig. \ref{fig:Test_Profile}a). The average velocity was found by fitting a piece-wise-linear ramp function to the extension data, the slope of which corresponds to the average velocity (Fig. \ref{fig:Test_Profile}b). The average velocity is then normalized by the pressurized rest length of the actuator to obtained the strain rate. The starting force ($F_0$) was calculated as the mean force during the two seconds prior to the start of the ramp. The peak force was taken as the force value when the extension first reached its target point. This was to avoid artifacts introduced by extension overshoot, which occurred at higher velocities. The peak force was then normalized by the starting force. 

These experimental force-velocity curves were then used to obtain model parameters for the McKibben actuators and viscoelastic sheaths as functions of pressure. For all experiments, the values of $d\varepsilon$, $\varepsilon_0$, $\varepsilon$, and $\hat v$ in the model were calculated relative to the pressurized rest length of the actuator. First, for each control McKibben actuator at each pressure, $\kappa_c$ and $\gamma_c$ were fitted using a nonlinear least squares method (code generated by using MATLAB Curve Fitting Toolbox). Parameter initialization was chosen based on the Equations~\ref{eqn:horizontalasymptote} and \ref{eqn:peakvelocity} for the horizontal asymptote and $\hat v_\alpha$. Specifically, the normalized force from the $\pm 10\text{ mm/s}$ tests were used as $FV_c(\hat v_\infty)$ to approximate $\kappa_c$, and the data from the $\pm 4\text{ mm/s}$ was used as the $\alpha$ point to estimate $\gamma_c$. To fit $\kappa_s$, $\gamma_s$, and $\beta_s$ for each material, diameter, and pressure, the corresponding McKibben parameters were set as $\kappa_c$ and $\gamma_c$ and not optimized. The parameter $\beta_s$ was initialized to 1, and $\kappa_s$ and $\gamma_s$ were initialized following the same procedure as in the plain McKibben case. The same optimization process was then carried out for these three parameters. 

\section{Results and Discussion}




\subsection{Characterization}
The force-velocity response of the twelve actuators was measured as a function of the pressure, sheath material, and mesh diameter (Fig.~\ref{fig:Experimental_Data}). In these figures, the normalized shortening velocity (negative of the extension strain rate) is reported in alignment with standard muscle physiology experiments. The shortening velocity is normalized by the pressurized rest length of the actuator (units of shortening velocity here are [1/s]). Common force-velocity features were found across all actuators. Unlike what is predicted in the model, all actuators showed an asymmetric force-velocity response, with a larger magnitude asymptote for extensions (negative shortening velocities) than for shortening. This reflects the nonlinear stiffness properties of the McKibben actuators that have been previously reported, with stiffness increasing with increased length \cite{Tondu2012}. Additionally, an increase in pressure led to a decrease in the height of force-velocity curve at all velocities and diameters, suggesting a more elastically dominant behavior at high pressures (Fig. \ref{fig:Parameter_Sweep}b,e). However, the difference in height for a given pressure increase diminished with increasing pressure. This is most pronounced in the 10 and 12 mm diameter actuators at the 5 psi level. This could be related to changes in the contact state of the inner bladder. In these larger actuators, the bladder is not in full contact with mesh at lower pressures, but at higher pressures has made full contact. This low-pressure discrepancy being related to the contact state is also supported by the observation this discrepancy is not seen in the 6 mm actuators, where the mesh is in full contact with the actuator even at low pressures. However, the mechanism that causes this contact state to result in a more viscous-dominated response would require additional investigation.

The addition of the viscoelastic polymer sheaths was successful in altering the force-velocity response of the McKibben actuators. In the case of the 6 and 12 mm diameter urethane actuators, a more viscous-dominating response was achieved, with the height of the force-velocity curve being higher than the control actuator response at all pressures and velocities. The 10 mm urethane actuator showed a different response, with the height in much closer agreement with the control. This could be due to the 10 mm urethane actuator requiring a larger diameter sheath than the other 10 mm actuators. The larger diameter sheath was used because the smaller sheath diameter consistently ruptured at higher pressures. However, this meant that the sheath was less in contact with the underlying McKibben than in the 6 and 12 mm cases and was thus less engaged. Conversely, the Ecoflex sheath led to a decrease in the height of the force-velocity curve in extension for all diameters and pressure, showing a more elastic-dominant response. In shortening, the Ecoflex actuators showed closer agreement with the control actuators. Finally, the effect of the Carbopol actuators varied. For the 6 mm actuator, almost no change was seem from the control actuator. In the 10 mm actuator, the response was much closer to that of the Ecoflex, showing a more elastic-dominant response. In the 12 mm case, the effect varied with pressure and direction of motion, with an increased height seen at 10 and 15 psi in extension, but no difference seen at 20 psi or in shortening at 10, 15 or 20 psi. 

This characterization is limited in a number of ways. The force-velocity curve, while relevant to the actuator in terms of its role as a model of skeletal muscle, is only one metric by which to determine these actuators' dynamic properties or the ability of these material sheaths to tune them. More complete characterization will require cyclic testing at various speeds to determine hysteresis as a function of velocity. Additionally, the minimum extension rate of 2 mm/s was near the horizontal asymptote for many actuators, resulting in poor characterization of the high slope region of the force-velocity curve near $\hat v=0$. A more complete investigation of the force-velocity curve will require lower velocities to be incorporated. These higher test rates also resulted in extension and shortening overshoot in the tests, which made the calculation of the peak force and the following force decay more challenging. These overshoots would be minimized with lower velocity tests. 

\subsection{Modeling}
\begin{figure}
    \centering
    \includegraphics[width=0.95\linewidth]{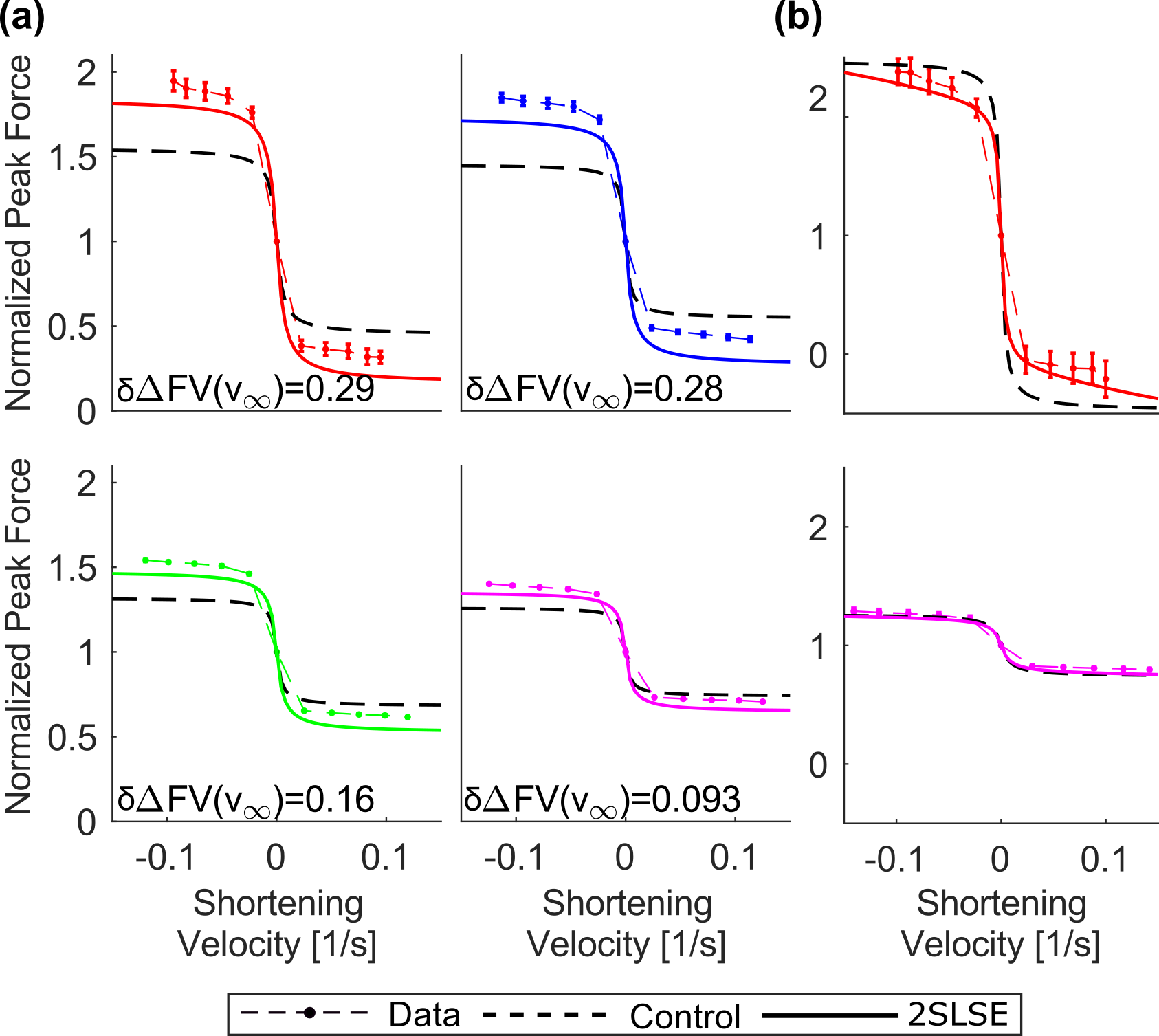}
    \caption{Comparison of 1 SLSE and 2 SLSE Model. For (a) and (b), the black dashed line shows the model fit of the corresponding plain Mckibben actuators (1 SLSE model), and the solid colored line shows the adjusted 2 SLSE model. As with Fig. \ref{fig:Experimental_Data}, the experimental data are shown as mean $\pm$ 1 STD. (a) 6mm Urethane VMA. For all pressures, the viscous nature of the urethane led to an increased height of the force-velocity curve, captured by the 2 SLSE model have a higher asymptote. As pressure increases and the McKibben stiffens, this asymptote difference decreases as the McKibben begins to dominate. ($\delta \Delta FV(\hat v_\infty) = FV_{2SLSE}(\hat v_\infty) - FV_{1SLSE}(\hat v_\infty)$). (b) 10mm Ecoflex VMA. At low pressure, the low viscous effects (large $\gamma_E$) of the Ecoflex sheath are able to change the slope of the force-velocity curve, but at higher pressures, the relative stiffness of the McKibben actuator again dominates, making the VMA response into alignment with the standard McKibben actuator.  }
    \label{fig:Model_Comparison}
    \vspace{-15 pt}
\end{figure}

The presented model was able to successfully capture major trends in the experimental force-velocity curves ($R^2 = 0.94\pm0.05$ for all actuators and pressures), and the changes in the VMA curve relative to the control curves can be explained through the model parameters. For example, in all actuators, an increase in pressure leads to a decrease in the height of the force-velocity curve. This is expected under this model, as increasing the pressure of the McKibben actuator increases its stiffness \cite{Tondu2012}, and this increased stiffness results in a lower $\kappa_c$ and thus $FV(\hat v_\infty)$. This model can also be used to explain changes in the force-velocity curves associated with the material sheaths (Fig. \ref{fig:Model_Comparison}). Based on preliminary materials testing, the urethane sheath would fall into the viscous dominant, long time constant class, and Ecoflex would fall into the elastic dominant, short time constant class (relative to the McKibben actuator). Therefore, we would expect that the urethane would cause an increase in the height of the force-velocity curve (Fig. \ref{fig:Parameter_Sweep}e). However, with increased pressure, the relative stiffness of the McKibben to the urethane sheath increase (decreasing $\beta_s$), so we would expect this difference to decrease with increased pressure as the weighted average begins to favor the McKibben actuator (Fig. \ref{fig:Model_Comparison}a). For the Ecoflex sheath, the relatively shorter time constant would lead to a high slope of the force-velocity curve, which is seen at low pressures (Fig. \ref{fig:Model_Comparison}b). However, as with the urethane sheath, an increased pressure leads to the McKibben properties dominating once again. 

While this model can capture many of the trends in the data, there are some limitations in its accuracy and predictive power. Both the McKibben actuators and the sheath materials are non-linearly elastic, with their stiffness increasing with increased strain. This results in an asymmetrical force-velocity curve, with a larger response for extension (negative shortening velocity) relative to shortening at the same rate. This cannot be captured by the linear springs in the proposed model. As a consequence, the model fits tend to under-predict extension responses and to over-predict shortening responses. Furthermore, the asymmetry also results in high parameter uncertainty. Improvements can be made through the inclusion of nonlinear spring elements and more appropriate models of the McKibben actuator at the cost of decreased interpretability of the model parameters. Additionally, the optimized sheath parameters tend to vary with pressure, whereas it would be expected that they would be pressure-independent for linear materials. However, a pressure dependence would be expected for nonlinear materials, as the McKibben actuator's pressure will determine the deformation state of the sheath material. In the future, this pressure dependence could be incorporated into the model as well, but it would require 3D geometric information about the actuator. Both of these issues could be addressed by incorporating a more complete quasi-static McKibben model \cite{Tondu2012,Al-Ibadi2017,Connollyetal2016trajectory} to capture the strain stiffening of the McKibben actuators and provide the geometry needed to estimate the sheath stiffness pressure dependence. 

Finally, this model only includes damping from standard dash-pot elements.  However,  previous work has shown that a velocity dependence in McKibben actuators can actually come from non-linear friction interactions in the mesh material\cite{Tondu2006} and Coulomb friction between the bladder material and the sheath \cite{Chou1994,Chou1996}. Future model development should incorporate such friction into a more complete model of the McKibben to replace one of the SLSEs in this model. 

\section{Conclusion and Future Works}
This work presents the characterization and modeling of the force-velocity relationships of viscoelastic McKibben actuators (VMA). These VMAs consist of a standard McKibben actuator surrounded by a viscoelastic polymer sheath. Iso-velocity experiments were performed to measure the force-velocity response of these actuators, and a simplified 1D model was developed to relate the shape of these experimental force-velocity curves to the material properties of the actuators. Using these polymer sheaths, we were able to successfully augment the force-velocity response of a standard McKibben, changing either its asymptotic height or its slope. The 1D model performed well in capturing the trends in these force-velocity curves, but missed key features, including the asymmetry in extension/shortening and the pressure dependence of sheath properties.

Future works on these actuators will include iso-velocity tests at slower speeds to further investigate the steep portion of the force-velocity curve near the $\hat v=0$ discontinuity. Additionally, to increase the predictive power of the model, more accurate quasistatic models of the McKibben's length-pressure-force properties will be implemented to replace the linear spring element. This will also require the measurement of the actuator geometry during quasi-static testing. Geometric information from these models will be used to capture the deformation-dependent properties of the sheath materials as well. With better predictive power, these models can be used as a design tool for creating actuators with a desired force-velocity response. Future characterization will also include cyclic testing of the actuators at various speeds to more robustly investigate their dynamic properties. The work presented here lays the foundation for the fabrication and design of pneumatic actuators with tunable force-velocity dynamics for broad applications in bioinspired and biomimetic robotics.

\section*{Acknowledgements}
This work was supported in part by the National Science Foundation (NSF) through grant no. FRR-2138873, and in part by NSF DBI-2015317 as part of the NSF/CIHR/DFG/FRQ/UKRI-MRC Next Generation Networks for Neuroscience Program. Any opinions, findings, and conclusions expressed in this material are those of the authors and do not necessarily reflect the views of the NSF.

\bibliographystyle{IEEEtran}
\bibliography{References}

\begin{thebibliography}{10}
\providecommand{\url}[1]{#1}
\csname url@samestyle\endcsname
\providecommand{\newblock}{\relax}
\providecommand{\bibinfo}[2]{#2}
\providecommand{\BIBentrySTDinterwordspacing}{\spaceskip=0pt\relax}
\providecommand{\BIBentryALTinterwordstretchfactor}{4}
\providecommand{\BIBentryALTinterwordspacing}{\spaceskip=\fontdimen2\font plus
\BIBentryALTinterwordstretchfactor\fontdimen3\font minus
  \fontdimen4\font\relax}
\providecommand{\BIBforeignlanguage}[2]{{%
\expandafter\ifx\csname l@#1\endcsname\relax
\typeout{** WARNING: IEEEtran.bst: No hyphenation pattern has been}%
\typeout{** loaded for the language `#1'. Using the pattern for}%
\typeout{** the default language instead.}%
\else
\language=\csname l@#1\endcsname
\fi
#2}}
\providecommand{\BIBdecl}{\relax}
\BIBdecl

\bibitem{daerden2002pneumatic}
F.~Daerden, D.~Lefeber \emph{et~al.}, ``Pneumatic artificial muscles: actuators
  for robotics and automation,'' \emph{European journal of mechanical and
  environmental engineering}, vol.~47, no.~1, pp. 11--21, 2002.

\bibitem{Hawkesetal2021Questions}
E.~Hawkes, C.~Majidi, and M.~Tolley, ``Hard questions for soft robotics,''
  \emph{Science Robotics}, vol.~6, no.~53, 2021.

\bibitem{Tondu2012}
B.~Tondu, ``{Modelling of the McKibben artificial muscle: A review},''
  \emph{Journal of Intelligent Material Systems and Structures}, vol.~23,
  no.~3, pp. 225--253, 2012.

\bibitem{Chou1996}
C.-P. Chou and B.~Hannaford, ``Measurement and modeling of mckibben pneumatic
  artificial muscles,'' \emph{IEEE Transactions on Robotics and Automation},
  pp. 90--102, 1996.

\bibitem{Delson2005}
N.~Delson, T.~Hanak, K.~Loewke, and D.~N. Miller, ``{Modeling and
  implementation of McKibben actuators for a hopping robot},'' \emph{2005
  International Conference on Advanced Robotics, ICAR '05, Proceedings}, vol.
  2005, pp. 833--840, 2005.

\bibitem{Kurumaya2016}
S.~Kurumaya, K.~Suzumori, H.~Nabae, and S.~Wakimoto, ``{Musculoskeletal
  lower-limb robot driven by multifilament muscles},'' \emph{ROBOMECH Journal},
  vol.~3, no.~1, pp. 1--15, 2016.

\bibitem{controlMcKibbenDesign}
K.~Dai, R.~Sukhnandan, M.~Bennington, K.~Whirley, R.~Bao, L.~Li, J.~P. Gill,
  H.~J. Chiel, and V.~A. Webster-Wood, ``Slugbot, an aplysia-inspired robotic
  grasper for studying control,'' \emph{Living Machines}, 2022.

\bibitem{Faudzi2018}
A.~A. Faudzi, N.~I. Azmi, M.~Sayahkarajy, W.~L. Xuan, and K.~Suzumori, ``{Soft
  manipulator using thin McKibben actuator},'' \emph{IEEE/ASME International
  Conference on Advanced Intelligent Mechatronics, AIM}, vol. 2018-July, pp.
  334--339, 2018.

\bibitem{Connollyetal2016trajectory}
F.~Connolly, C.~Walsh, and K.~Bertoldi, ``Automatic design of fiber-reinforced
  soft actuators for trajectory matching,'' \emph{Proceedings of the National
  Academy of Sciences}, vol. 114, no.~1, pp. 51--56, 2016.

\bibitem{Tschiersky2020}
M.~Tschiersky, E.~E. Hekman, D.~M. Brouwer, J.~L. Herder, and K.~Suzumori, ``{A
  Compact McKibben Muscle Based Bending Actuator for Close-to-Body Application
  in Assistive Wearable Robots},'' \emph{IEEE Robotics and Automation Letters},
  vol.~5, no.~2, pp. 3042--3049, 2020.

\bibitem{Koizumi2020}
S.~Koizumi, T.~H. Chang, H.~Nabae, G.~Endo, K.~Suzumori, M.~Mita, K.~Saitoh,
  K.~Hatakeyama, S.~Chida, and Y.~Shimada, ``{Soft Robotic Gloves with Thin
  McKibben Muscles for Hand Assist and Rehabilitation},'' \emph{Proceedings of
  the 2020 IEEE/SICE International Symposium on System Integration, SII 2020},
  pp. 93--98, 2020.

\bibitem{Rosaliaetal2022sleeve}
L.~Rosalia, C.~Ozturk, J.~Coll-Font, Y.~Fan, Y.~Nagata, M.~Singh, D.~Goswami,
  A.~Mauskapf, S.~Chen, R.~A. Eder, E.~M. Goffer, J.~H. Kim, S.~Yurista, B.~P.
  Bonner, A.~N. Foster, R.~A. Levine, E.~R. Edelman, M.~Panagia, J.~L.
  Guerrero, E.~T. Roche, and C.~T. Nguyen, ``A soft robotic sleeve mimicking
  the haemodynamics and biomechanics of left ventricular pressure overload and
  aortic stenosis,'' \emph{Nature Biomedical Engineering}, vol.~6, pp.
  1134--1147, 2022.

\bibitem{Parketal2012}
Y.-L. Park, B.-r. Chen, C.~Majidi, R.~J. Wood, R.~Nagpal, and E.~Goldfield,
  ``Active modular elastomer sleeve for soft wearable assistance robots,'' in
  \emph{2012 IEEE/RSJ International Conference on Intelligent Robots and
  Systems}, 2012, pp. 1595--1602.

\bibitem{Tondu2006}
B.~Tondu and S.~D. Zagal, ``{McKibben artificial muscle can be adapted to be in
  accordance with the Hill skeletal muscle model},'' \emph{Proceedings of the
  First IEEE/RAS-EMBS International Conference on Biomedical Robotics and
  Biomechatronics, 2006, BioRob 2006}, vol. 2006, no.~3, pp. 714--720, 2006.

\bibitem{Klute1999}
G.~K. Klute, J.~M. Czerniecki, and B.~Hannaford, ``{McKibben artificial
  muscles: Pneumatic actuators with biomechanical intelligence},''
  \emph{IEEE/ASME International Conference on Advanced Intelligent
  Mechatronics, AIM}, pp. 221--226, 1999.

\bibitem{Gollobetal2022Joint}
S.~Gollob, J.~Poss, G.~Memoli, and E.~Roche, ``A multi-material,
  anthropomorphic metacarpophalangeal joint with abduction and adduction
  actuated by soft artificial muscles,'' \emph{IEEE Robotics and Automation
  Letters}, vol.~7, no.~3, pp. 5882--5887, 2022.

\bibitem{Al-Ibadi2017}
A.~Al-Ibadi, S.~Nefti-Meziani, and S.~Davis, ``{Efficient structure-based
  models for the McKibben contraction pneumatic muscle actuator: The full
  description of the behaviour of the contraction PMA},'' \emph{Actuators},
  vol.~6, no.~4, 2017.

\bibitem{Olsenetal2022McKibbenModeling}
G.~Olsen, H.~Manjarrez, J.~Adams, and Y.~Meng\"{u}\c{c}, ``Experimentally
  identified models of mckibben soft actuators as primary movers and passive
  structures,'' \emph{Journal of Mechanisms and Robotics}, vol.~14, no.
  JMR-20-1425, pp. 011\,006--1--011\,006--15, 2022.

\bibitem{Kothera2009}
C.~S. Kothera, M.~Jangid, J.~Sirohi, and N.~M. Wereley, ``{Experimental
  characterization and static modeling of McKibben actuators},'' \emph{Journal
  of Mechanical Design, Transactions of the ASME}, vol. 131, no.~9, pp.
  0\,910\,101--09\,101\,010, 2009.

\bibitem{Taimooretal2019}
T.~Hassan, M.~Cianchetti, M.~Moatamedi, B.~Mazzolai, C.~Laschi, and P.~Dario,
  ``Finite-element modeling and design of a pneumatic braided muscle actuator
  with multifunctional capabilities,'' \emph{IEEE/ASME Transactions on
  Mechatronics}, vol.~24, no.~1, pp. 109--119, 2019.

\bibitem{Chou1994}
C.~P. Chou and B.~Hannaford, ``{Static and dynamic characteristics of McKibben
  pneumatic artificial muscles},'' \emph{Proceedings - IEEE International
  Conference on Robotics and Automation}, no. pt 1, pp. 281--286, 1994.

\bibitem{Yu1999}
S.~N. Yu, P.~E. Crago, and H.~J. Chiel, ``{Biomechanical properties and a
  kinetic simulation model of the smooth muscle I2 in the buccal mass of
  Aplysia},'' \emph{Biological Cybernetics}, vol.~81, no. 5-6, pp. 505--513,
  1999.

\bibitem{Carbopol1}
T.~J. Hinton, A.~Hudson, K.~Pusch, A.~Lee, and A.~W. Fienberg, ``3d printing
  pdms elastomer in a hydrophilic support bath via freeform reversible
  embedding,'' \emph{ACS Biomaterials Science \& Engineering}, 2016.

\end{thebibliography}

\newpage

\end{document}